\documentclass[pdflatex,sn-mathphys]{sn-jnl}

\usepackage{markdown}

\jyear{2022}%
 
\theoremstyle{thmstyleone}%
\theoremstyle{thmstyletwo}%

\theoremstyle{thmstylethree}%

\unnumbered% uncomment this for unnumbered level heads

\begin{document}

\title[\textit{Needs}-aware Artificial Intelligence]{\textit{Needs}-aware Artificial Intelligence:\linebreak 
AI that `serves [human] needs'}

\author[1,*]{\fnm{Ryan} \sur{Watkins}}\email{rwatkins@gwu.edu}

\author[2,3]{\fnm{Soheil} \sur{Human}}\email{soheil.human@wu.ac.at}

%\equalcont{These authors contributed equally to this work.}
%\author[1,2]{\fnm{Third} \sur{Author}}\email{iiiauthor@gmail.com}
%\equalcont{These authors contributed equally to this work.}

\affil[1]{\orgname{George Washington University}, \orgaddress{\street{G Street NW}, \city{Washington}, \postcode{20052}, \state{DC}, \country{USA}}}
\affil[*]{Corresponding author}

\affil[2]{\orgdiv{Sustainable Computing Lab, Institute for Information Systems and New Media}, \orgname{Vienna University of Economics and Business}, \orgaddress{\street{Welthandelsplatz 1}, \city{Vienna}, \postcode{A-1020}, \country{Austria, EU}}}

\affil[3]{\orgdiv{Department of Philosophy \& Vienna Cognitive Science Hub}, \orgname{University of Vienna}, \orgaddress{\street{Universitätsstraße 7}, \city{Vienna}, \postcode{A-1010}, \country{Austria, EU}}}

\maketitle

\section{Keywords}
needs, needs-aware, sociotechnical, interdisciplinary

\section{Abstract}
By defining the current limits (and thereby the frontiers), many boundaries are shaping, and will continue to shape, the future of Artificial Intelligence (AI). We push on these boundaries in order to make further progress into what were yesterday's frontiers. They are both pliable and resilient---always creating new boundaries of what AI can (or should) achieve. Among these are technical boundaries (such as processing capacity), psychological boundaries (such as human trust in AI systems), ethical boundaries (such as with AI weapons), and conceptual boundaries (such as the AI people can imagine).  It is within this final category\footnote{while it can play a fundamental role in all other boundaries} that we find the construct of \textit{needs} and the limitations that our current concept of \textit{need} places on the future AI. 

\section{Serve [Human] Needs}

Multiple AI advocates (including Kai-Fu Lee \cite{lee_oreilly_2021} and Ben Shneiderman \cite{Shneiderman2020HCAIgoals, shneiderman2022human}), among many others, have posited that a primary goal of AI (and Human-centric AI\footnote{HCAI}) is to serve human \textit{needs}.  A laudable goal for sure, but there is a great deal of history, controversy, and complexity packed into both the word \textit{need} and the overarching construct of \textit{needs} \cite{humanOntologyRepresentingHuman2017a}.  Thus, if serving \textit{needs} is to remain an ambition of our AI systems, further attention (i.e., dialogue, research, guidelines, policies) and collaboration across multiple disciplines is required to develop the construct of \textit{needs} into a pragmatic tool that be applied to shape the very goals of what future AI can and should achieve.  

\textit{Need} is a commonplace word (such as, “I \textit{need} coffee”), making it easy to overlook that the term has specific meaning, definition, connotation, and power. Its power, for example, stems from the connotation that the object of the statement (such as, coffee in the example above) seems to be absolutely necessary and without alternative.
In other words, coffee is required to satisfy the implied \textit{need}.
Coffee may not be sufficient, but tea or water alone definitely won't do.\footnote{A useful exercise can be to go a day, or week, without using the word ``need'' at all; quickly allowing each of us to recognize just how often we use the power of the term in our daily activities.}  

Most of us routinely leverage this power (as do politicians and advertisers) when we use the word \textit{need} to effectively eliminate other options (such as, ``Cryptocurrency companies \textit{need} national regulations'', when, e.g., international regulations, market-based instruments, co-regulation, self-regulation, education \cite{regulationoecd}, and end-user empowerment \cite{humanEnduserEmpowermentInterdisciplinary2020} might be other viable options to be considered)\footnote{Here, we suspend our judgment regarding the national regulations of cryptocurrency companies since this is out of the scope of this article; the point here is that by using ``need'' we imply necessity [without evidence] and infer that any action \textit{must} include national regulations when other options should also be considered.}. 
We do this because \textit{need} statements typically induce the desired associated behaviors (such as, choosing national regulations rather than other alternatives); though typically creating ethical difficulties both for those defining the \textit{need}, and those tasked with satisfying the \textit{need}.  
Defining \textit{needs}, after all, is not just about an academic concept; rather it can determine whose \textit{needs} are prioritized, who gets resources and who does not, and how inequalities are considered in meeting the basics of the human condition.
In these cases, \textit{need} is a very powerful construct---and yet it remains one that we have little understanding of or agreement on. 
Those who define \textit{needs} (whether they be individuals for themselves or for others, institutions such as companies or governments, or in the future AI systems) have both implicit and explicit power---and yet we rarely recognize that power since it is routinely lost in the common usage of the term.
If an AI system, for example, were permitted to determine [and prioritize] a patient's \textit{needs} [and the satisfiers of that needs], the power of the tool is substantially greater than if it only offers options for medical care.

It is worth emphasizing that being \textit{in need} (and accordingly serving \textit{needs}) is not limited to individual humans. \textit{Needs} can be associated with different types of systems (e.g. life forms, organizations, societies). Therefore, \textit{needs}-aware AI systems \cite{HumanWatkins2022NeedsAI} should ideally consider different systems' needs (plural) on different levels and different contexts \textit{sustainably}.

\section{What Are Needs?}

Distinguishing between what is necessary (i.e., \textit{needs}) and what is desired (i.e., transitory wants, cravings, motivators) has multiple ethical implications for AI and AI developers. This distinction is easily lost, for example, when put into the context of determining what potential clients or customers will purchase (where people might elect to spend their own money on what they desire over what is necessary).  While ascertaining peoples' desires is not always an easy task, it is relatively much easier than identifying and prioritizing their \textit{needs} (i.e., the goal of a needs assessment \cite{watkins2012guide}). Different scholars, such as the philosopher Stephen McLeod, have even questioned if people are capable of knowing their \textit{needs} at all \cite{mcleod2011knowledge}.  

For AI developers, for instance, the challenges of this distinction (i.e., \textit{needs} from wants)\footnote{Though we recognize that colleagues in multiple disciplines have also proposed typologies for ``needs'', we will not address those in this article. Typologies are one of many topics we hope will be taken up in future interdisciplinary dialogues/debates.} leads to an ethical difficulty that spans the continuum stretching from creating systems that merely meet consumers stated desires at the moment, to systems that assist in resolving [human] \textit{needs} even when people may be unaware of the benefits at the time. Moving from basic perspectives of \textit{needs} (e.g., \textit{needs} are what people say they \textit{need}, or \textit{needs} are only what motivates an individual to take action \cite{maslow1943theory}) to a more robust and multidimensional definition and understanding of \textit{needs} (e.g., \textit{needs} are gaps between desired accomplishments and current achievements at multiple interdependent levels \cite{watkins2012guide}) brings many benefits, but also introduces complexity for AI developers creating (or co-creating) Sustainable \textit{H}uman-centric, \textit{A}ccountable, \textit{L}awful, and \textit{E}thical AI (Sustainable HALE AI \cite{humanHALEWHALEFramework2022}) systems (for instance, balancing individual, organizational, and societal \textit{needs} that are routinely in conflict).

What are \textit{needs}? What are not \textit{needs}?  How do we prioritize among \textit{needs}? How do my \textit{needs} relate to your \textit{needs}, and how do our \textit{needs} relate to the \textit{needs} of others? How do we measure \textit{needs}? How can we utilize needs? What will satisfy a \textit{need}, and how will we know if the \textit{need} has been satisfied? How can AI serve \textit{needs} and still be economically viable? How can/will different sociopolitical, socio-economic, socio-technical and socio-cognitive aspects influence the co-creation of \textit{needs-aware AI} systems, and how can/will such aspects be appropriately considered in a \textit{Sustainable HALE} co-creation of such systems? These, and many other, questions have been and are still debated within and across multiple disciplines (e.g., philosophy, ethics, law, social work, education, business, economics, political science, sociology, management, cognitive science, psychology, and engineering).  These debates have not, however, reached a resolution; and we suggest that this does, and will continue to, create pragmatic boundaries on what AI can and should achieve.
Likewise, without answers to these questions (or at least many/most of them) it might be ethically challenging to ask (or expect) AI developers (or AI systems) to assess the \textit{needs} of others, and then to use the results of those assessments to create AI systems that meet ethical standards. 

\section{Roles for \textit{Needs}}

AI developers are often placed in a so-called \textit{social dilemmas}---with societal good on one side and commercial pressures on the other \cite{struemke2021social}.
Part of the solution to these dilemmas (beyond ethical, legal, and regulatory frameworks) could be the introduction of well-defined and measurable \textit{needs}\footnote{Calling for well-define and measurable \textit{needs} (or needs satisfaction) does not mean that we are advocating absolutist perspectives on \textit{needs}. With that in mind, we propose that, among others, considering \textit{disagreements} \cite{humanSupportingPluralismArtificial2018a} should be an important aspect of \textit{needs-aware} AI systems (see \cite{HumanWatkins2022NeedsAI} for a more detailed discussion on \textit{measuring, \textit{explicitizing}, \textit{utilizing}, or \textit{enactizing} needs}).}.
For example, by identifying and measuring \textit{needs} (i.e., societal, organizational, and individual \textit{needs}) we can contribute to building the foundations for finding an appropriate equilibrium that serves \textit{needs} in meaningful and balanced ways; while providing tools capable of guiding AI ethics.  As an integrated component of \textit{H}uman-centric, \textit{A}ccountable, \textit{L}awful, and \textit{E}thical AI (or HALE AI) \cite{humanHALEWHALEFramework2022}, the construct of \textit{needs} can, we suggest, add value and push the boundaries of AI development from chasing wants, to serving \textit{needs}\footnote{Considering that meeting different systems' interrelated (and sometimes conflicting) needs in a sustainable manner is crucially important for our societies, re-thinking needs (and needs satisfaction) into AI can not only contribute toward the development of HALE AI but \textit{Sustainable} HALE AI \cite{humanHALEWHALEFramework2022}.}.

\textit{Needs} can thereby contribute in multiple roles in the development of AI. HCAI developers, for example, can utilize \textit{needs} to identify and prioritize both what the systems can and should achieve; meeting peoples' desires and also serving their \textit{needs}.  AI systems, for instance, can use measurable \textit{needs} to evaluate their own performance in resolving \textit{needs}, while at the same time assisting people in making decisions where the complex relationships among \textit{needs} must be weighed.  Meanwhile, policymakers can utilize well-defined societal \textit{needs} to craft effective policy, regulatory, and ethical frameworks.  As such, precise, comprehensive, and transparent constructs of \textit{needs} can play many vital roles in the future development of AI (and our digital societies).  

\section{What Next?}
If AI is going to serve our \textit{needs}, then we have to answer some of these questions, and discover new questions that are waiting below the surface. From our perspective this is an urgent matter since these questions will not be answered quickly and without debate, and AI researchers and developers must be part of the professional dialogues in order for useful guidance to be achieved.
No single discipline or field can come to resolution on these matters, and thereby \textit{needs} are illustrative of the types of broad interdisciplinary challenges (bringing together STEM, social science, and humanities scholars and practitioners) that will be the hallmark of future decades of AI research and development.
At the same time, the development of new AI systems will not necessarily wait for academic debates---as history shows.

\textit{Needs}, both as a construct and professional term, can (and should) be a fundamental element of ethical (and sociotechnical) frameworks and the tools that are derived from those frameworks.
We must use the word with the same precision and with the same care as we accord to terms such as ``values'' or ``rights''.
We must also work to create a shared understanding of what \textit{needs} are, defining them in manners that can transcend disciplinary boundaries and allow us to align individual, organizational, and societal \textit{needs} \cite{kaufman2019alignment}.

If we give up, however, and choose not to become precise in our construct of \textit{need} (our language when discussing \textit{needs}), and the operational definitions required for future \textit{Needs-aware AI} systems, then we will be left with AI that merely helps us meet our transitory wants, desires, cravings, motivations, or passions\footnote{and maybe only as a by-product some of our ``needs'', though we would have a hard time knowing it.}.
All of which may be profitable and favorable at times, but none of which are sufficient (nor necessary) for meeting our ideal of future AI that has the capacity to serve [human] \textit{needs}.

The path to \textit{needs-aware} AI will take time. Truly interdisciplinary dialogue and collaboration requires time.\footnote{while developers might not wait for it.} 
From philosophy to computer science, and cognitive science to social science, many disciplines have contributions to offer, and yet there is much to learn about those potential contributions as we prepare for the future.  
For instance, many scholars who study the psychology of need do not also follow current development in computer science and AI; and the reverse is true as well.  
We therefore suggest that the process of interdisciplinary collaboration on \textit{needs-aware AI} must begin soon, to ensure that the distinction of \textit{needs} isn't lost (or assumed) as technologies develop over the next decade(s).
This can begin here, with responses to this initial editorial; and then grow through cross-disciplinary dialogue. 
Whether it is maintaining \textit{needs} as a distinct concept in [re]presentations, high-lighting the unique role of \textit{needs} as systemic or algorithm features, or applying \textit{needs} in design and co-creation processes, the role of \textit{needs} in the future of AI depends on recognizing the power and value of this frequently misunderstood construct.

\section{Statements and Declarations}
No funding was received to assist with the preparation of this manuscript. The authors have no relevant financial or non-financial interests to disclose.

\bibliography{z__bibliography}% common bib file

%% BioMed_Central_Bib_Style_v1.01

\begin{thebibliography}{14}
% BibTex style file: bmc-mathphys.bst (version 2.1), 2014-07-24
\ifx \bisbn   \undefined \def \bisbn  #1{ISBN #1}\fi
\ifx \binits  \undefined \def \binits#1{#1}\fi
\ifx \bauthor  \undefined \def \bauthor#1{#1}\fi
\ifx \batitle  \undefined \def \batitle#1{#1}\fi
\ifx \bjtitle  \undefined \def \bjtitle#1{#1}\fi
\ifx \bvolume  \undefined \def \bvolume#1{\textbf{#1}}\fi
\ifx \byear  \undefined \def \byear#1{#1}\fi
\ifx \bissue  \undefined \def \bissue#1{#1}\fi
\ifx \bfpage  \undefined \def \bfpage#1{#1}\fi
\ifx \blpage  \undefined \def \blpage #1{#1}\fi
\ifx \burl  \undefined \def \burl#1{\textsf{#1}}\fi
\ifx \doiurl  \undefined \def \doiurl#1{\url{https://doi.org/#1}}\fi
\ifx \betal  \undefined \def \betal{\textit{et al.}}\fi
\ifx \binstitute  \undefined \def \binstitute#1{#1}\fi
\ifx \binstitutionaled  \undefined \def \binstitutionaled#1{#1}\fi
\ifx \bctitle  \undefined \def \bctitle#1{#1}\fi
\ifx \beditor  \undefined \def \beditor#1{#1}\fi
\ifx \bpublisher  \undefined \def \bpublisher#1{#1}\fi
\ifx \bbtitle  \undefined \def \bbtitle#1{#1}\fi
\ifx \bedition  \undefined \def \bedition#1{#1}\fi
\ifx \bseriesno  \undefined \def \bseriesno#1{#1}\fi
\ifx \blocation  \undefined \def \blocation#1{#1}\fi
\ifx \bsertitle  \undefined \def \bsertitle#1{#1}\fi
\ifx \bsnm \undefined \def \bsnm#1{#1}\fi
\ifx \bsuffix \undefined \def \bsuffix#1{#1}\fi
\ifx \bparticle \undefined \def \bparticle#1{#1}\fi
\ifx \barticle \undefined \def \barticle#1{#1}\fi
\bibcommenthead
\ifx \bconfdate \undefined \def \bconfdate #1{#1}\fi
\ifx \botherref \undefined \def \botherref #1{#1}\fi
\ifx \url \undefined \def \url#1{\textsf{#1}}\fi
\ifx \bchapter \undefined \def \bchapter#1{#1}\fi
\ifx \bbook \undefined \def \bbook#1{#1}\fi
\ifx \bcomment \undefined \def \bcomment#1{#1}\fi
\ifx \oauthor \undefined \def \oauthor#1{#1}\fi
\ifx \citeauthoryear \undefined \def \citeauthoryear#1{#1}\fi
\ifx \endbibitem  \undefined \def \endbibitem {}\fi
\ifx \bconflocation  \undefined \def \bconflocation#1{#1}\fi
\ifx \arxivurl  \undefined \def \arxivurl#1{\textsf{#1}}\fi
\csname PreBibitemsHook\endcsname

%%% 1
\bibitem{lee_oreilly_2021}
\begin{botherref}
\oauthor{\bsnm{Lee}, \binits{K.-F.}},
\oauthor{\bsnm{OReilly}, \binits{T.}}:
Meet the Expert: How AI Will Change Our World by 2041.
OReilly Media, Inc.
(2021).
\url{https://learning.oreilly.com/videos/meet-the-expert/0636920623939/0636920623939-video335577/}
\end{botherref}
\endbibitem

%%% 2
\bibitem{Shneiderman2020HCAIgoals}
\begin{barticle}
\bauthor{\bsnm{Shneiderman}, \binits{B.}}:
\batitle{Design lessons from ai’s two grand goals: Human emulation and useful
  applications}.
\bjtitle{IEEE Transactions on Technology and Society}
\bvolume{1}(\bissue{2}),
\bfpage{73}--\blpage{82}
(\byear{2020})
\end{barticle}
\endbibitem

%%% 3
\bibitem{shneiderman2022human}
\begin{bbook}
\bauthor{\bsnm{Shneiderman}, \binits{B.}}:
\bbtitle{Human-Centered AI}.
\bpublisher{Oxford University Press},
\blocation{Oxford, UK}
(\byear{2022})
\end{bbook}
\endbibitem

%%% 4
\bibitem{humanOntologyRepresentingHuman2017a}
\begin{bchapter}
\bauthor{\bsnm{Human}, \binits{S.}},
\bauthor{\bsnm{Fahrenbach}, \binits{F.}},
\bauthor{\bsnm{Kragulj}, \binits{F.}},
\bauthor{\bsnm{Savenkov}, \binits{V.}}:
\bctitle{Ontology for {{Representing Human Needs}}}.
In: \beditor{\bsnm{R{\'o}{\.z}ewski}, \binits{P.}},
\beditor{\bsnm{Lange}, \binits{C.}} (eds.)
\bbtitle{Knowledge {{Engineering}} and {{Semantic Web}}}.
\bsertitle{Communications in {{Computer}} and {{Information Science}}},
pp. \bfpage{195}--\blpage{210}.
\bpublisher{{Springer International Publishing}},
\blocation{{Cham}}
(\byear{2017})
\end{bchapter}
\endbibitem

%%% 5
\bibitem{regulationoecd}
\begin{botherref}
\oauthor{\bparticle{\uppercase{OECD}} \bsnm{Report}}:
Regulation, alternatives traditional.
\url{https://www.oecd.org/gov/regulatory-policy/42245468.pdf}
\end{botherref}
\endbibitem

%%% 6
\bibitem{humanEnduserEmpowermentInterdisciplinary2020}
\begin{bchapter}
\bauthor{\bsnm{Human}, \binits{S.}},
\bauthor{\bsnm{Gsenger}, \binits{R.}},
\bauthor{\bsnm{Neumann}, \binits{G.}}:
\bctitle{End-user empowerment: {{An}} interdisciplinary perspective}.
In: \bbtitle{Proceedings of the 53rd Hawaii International Conference on System
  Sciences},
\bconflocation{{Hawaii, United States}},
pp. \bfpage{4102}--\blpage{4111}
(\byear{2020})
\end{bchapter}
\endbibitem

%%% 7
\bibitem{HumanWatkins2022NeedsAI}
\begin{barticle}
\bauthor{\bsnm{Human}, \binits{S.}},
\bauthor{\bsnm{Watkins}, \binits{R.}}:
\batitle{{Needs} {and} {Artificial} {Intelligence}}.
\bjtitle{arXiv}
(\bissue{arXiv:2202.04977 [cs.AI]})
(\byear{2022}).
\doiurl{10.48550/arXiv.2202.04977}
\end{barticle}
\endbibitem

%%% 8
\bibitem{watkins2012guide}
\begin{bbook}
\bauthor{\bsnm{Watkins}, \binits{R.}},
\bauthor{\bsnm{Meiers}, \binits{M.W.}},
\bauthor{\bsnm{Visser}, \binits{Y.}}:
\bbtitle{A Guide to Assessing Needs: Essential Tools for Collecting
  Information, Making Decisions, and Achieving Development Results}.
\bpublisher{World Bank Publications},
\blocation{D.C., USA}
(\byear{2012})
\end{bbook}
\endbibitem

%%% 9
\bibitem{mcleod2011knowledge}
\begin{barticle}
\bauthor{\bsnm{McLeod}, \binits{S.K.}}:
\batitle{Knowledge of need}.
\bjtitle{International Journal of Philosophical Studies}
\bvolume{19}(\bissue{2}),
\bfpage{211}--\blpage{230}
(\byear{2011})
\end{barticle}
\endbibitem

%%% 10
\bibitem{maslow1943theory}
\begin{barticle}
\bauthor{\bsnm{Maslow}, \binits{A.H.}}:
\batitle{A theory of human motivation.}
\bjtitle{Psychological Review}
\bvolume{50}(\bissue{4}),
\bfpage{370}
(\byear{1943})
\end{barticle}
\endbibitem

%%% 11
\bibitem{humanHALEWHALEFramework2022}
\begin{botherref}
\oauthor{\bsnm{Human}, \binits{S.}}:
{{THE HALE WHALE}}: {{A Framework}} for the {{Co-creation}} of {{Sustainable}},
  {{Human-centric}}, {{Accountable}}, {{Lawful}}, and {{Ethical Digital
  Sociotechnical Systems}}.
Sustainable Computing Paper Series
(2022/01)
(2022)
\end{botherref}
\endbibitem

%%% 12
\bibitem{struemke2021social}
\begin{botherref}
\oauthor{\bsnm{Strümke}, \binits{I.}},
\oauthor{\bsnm{Slavkovik}, \binits{M.}},
\oauthor{\bsnm{Madai}, \binits{V.I.}}:
The Social Dilemma in Artificial Intelligence Development and Why We Have to
  Solve It
(2021)
\end{botherref}
\endbibitem

%%% 13
\bibitem{humanSupportingPluralismArtificial2018a}
\begin{bchapter}
\bauthor{\bsnm{Human}, \binits{S.}},
\bauthor{\bsnm{Bidabadi}, \binits{G.}},
\bauthor{\bsnm{Savenkov}, \binits{V.}}:
\bctitle{Supporting {{Pluralism}} by {{Artificial Intelligence}}:
  {{Conceptualizing Epistemic Disagreements As Digital Artifacts}}}.
\bsertitle{{{PT-AI}} 2017: {{Philosophy}} and {{Theory}} of {{Artificial
  Intelligence}} 2017},
pp. \bfpage{190}--\blpage{193}.
\bpublisher{{Springer and Springer}},
\blocation{{Leeds}}
(\byear{2018})
\end{bchapter}
\endbibitem

%%% 14
\bibitem{kaufman2019alignment}
\begin{barticle}
\bauthor{\bsnm{Kaufman}, \binits{R.}}:
\batitle{Alignment and success: Applying the hierarchy of planning and the
  needs-assesment hierarchy}.
\bjtitle{Performance Improvement}
\bvolume{58}(\bissue{7}),
\bfpage{24}--\blpage{28}
(\byear{2019})
\end{barticle}
\endbibitem

\end{thebibliography}
%% if required, the content of .bbl file can be included here once bbl is generated
%%\input sn-article.bbl

%% Default %%
%%\input sn-sample-bib.tex%

% Make a new page at the end of the PDF file
% \newpage
% Include the reviewer responses
% 
%\markdownInput{reviewer_responses.md}
% Comment the above two commands if you do not want to include the reviewer responses in the article file... use reviewer_responses.tex as the "main" file to produce the responses as a separate PDF file...
\end{document}